\pdfoutput=1

\documentclass[11pt]{article}

\usepackage[final]{acl}

\usepackage{times}
\usepackage{latexsym}

\usepackage[T1]{fontenc}

\usepackage[utf8]{inputenc}

\usepackage{microtype}

\usepackage{inconsolata}

\usepackage{graphicx}
\usepackage{wrapfig}
\usepackage{algorithm}
\usepackage{algorithmic}
\usepackage{amsmath}

\usepackage{enumitem}
\usepackage{etoolbox}
\usepackage{booktabs,makecell,tabularx} 
\usepackage{resizegather}
\usepackage{multirow}
\usepackage{amssymb}
\usepackage{amsmath}
\usepackage{graphicx}
\usepackage{tcolorbox}
\usepackage{subcaption}
\usepackage{colortbl}
\usepackage{listings}
\usepackage{xcolor}

\definecolor{deepgreen}{rgb}{0.0, 0.6, 0.0}
\title{Decoupled Proxy Alignment: Mitigating Language Prior Conflict for Multimodal Alignment in MLLMs}

\author{
    \textbf{Chenkun Tan},
    ~\textbf{Pengyu Wang},
    ~\textbf{Shaojun Zhou},
    ~\textbf{Botian Jiang}, \\
    ~\textbf{Zhaowei Li},
    ~\textbf{Dong Zhang},
    ~\textbf{Xinghao Wang},
    ~\textbf{Yaqian Zhou},
    ~\textbf{Xipeng Qiu} \thanks{Corresponding author.} \\
    Fudan University\\
    {\tt 	\{cktan25,pywang24,sjzhou24,btjiang23\}@m.fudan.edu.cn} \\
    {\tt 	\{zhouyaqian,xpqiu\}@fudan.edu.cn} \\
}

\begin{document}
\maketitle
\begin{abstract}
Multimodal large language models (MLLMs) have gained significant attention due to their impressive ability to integrate vision and language modalities. Recent advancements in MLLMs have primarily focused on improving performance through high-quality datasets, novel architectures, and optimized training strategies. However, in this paper, we identify a previously overlooked issue, \textbf{language prior conflict}, a mismatch between the inherent language priors of large language models (LLMs) and the language priors in training datasets. This conflict leads to suboptimal vision-language alignment, as MLLMs are prone to adapting to the language style of training samples. To address this issue, we propose a novel training method called \textbf{Decoupled Proxy Alignment (DPA)}. DPA introduces two key innovations: (1) the use of a proxy LLM during pretraining to decouple the vision-language alignment process from language prior interference, and (2) dynamic loss adjustment based on visual relevance to strengthen optimization signals for visually relevant tokens. Extensive experiments demonstrate that DPA significantly mitigates the language prior conflict, achieving superior alignment performance across diverse datasets, model families, and scales. Our method not only improves the effectiveness of MLLM training but also shows exceptional generalization capabilities, making it a robust approach for vision-language alignment. \footnote{Our code is available at \url{https://github.com/fnlp-vision/DPA}.}
\end{abstract}

\section{Introduction}
\begin{figure}[!t]
    \centering
    \subfloat[The Dataset Quality Paradox on CVBench.]{\includegraphics[width=0.88\columnwidth]{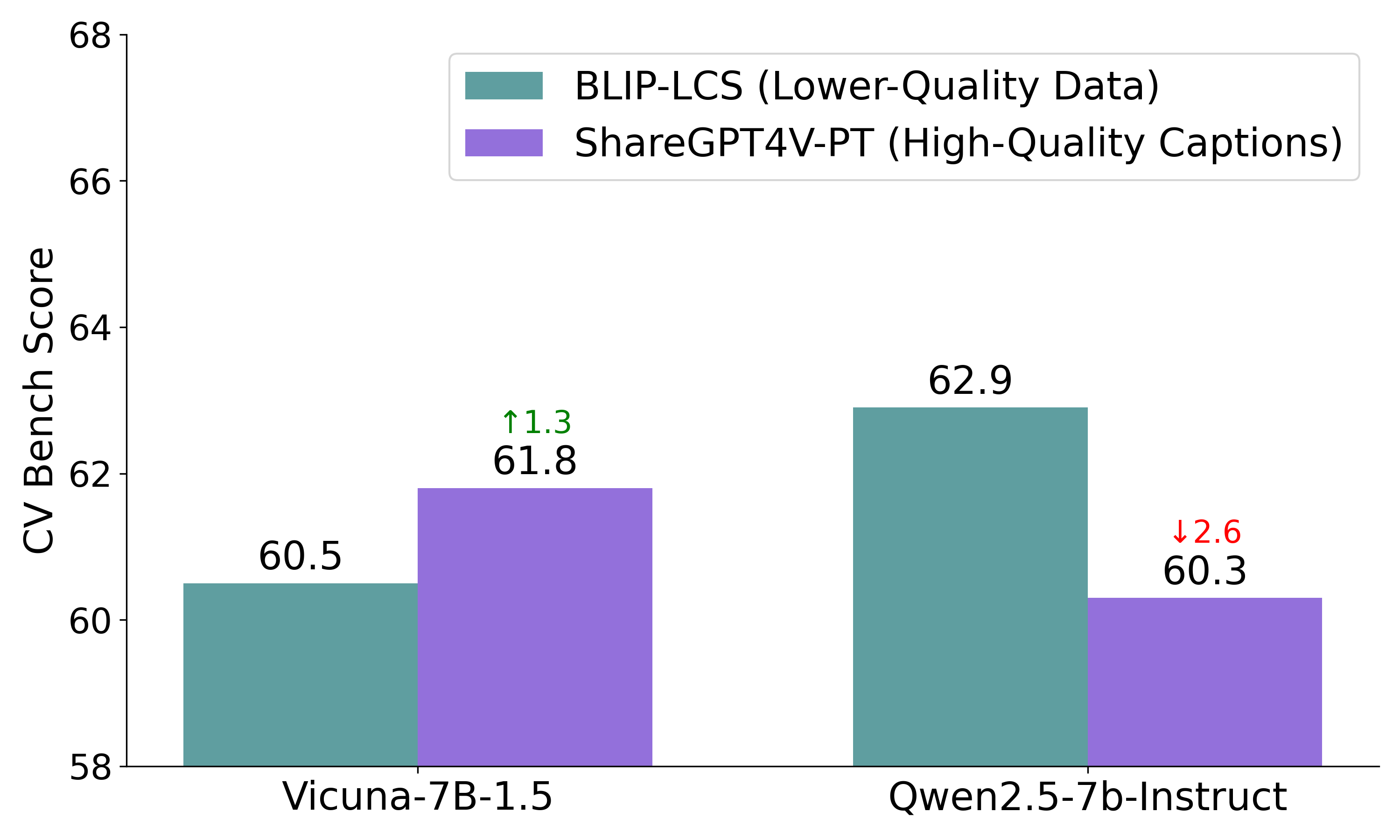}\label{fig:first_1}}\vspace{4pt}
    \subfloat[Loss change (\%) for linguistically relevant and visually relevant words before and after training.]{\includegraphics[width=0.88\columnwidth]{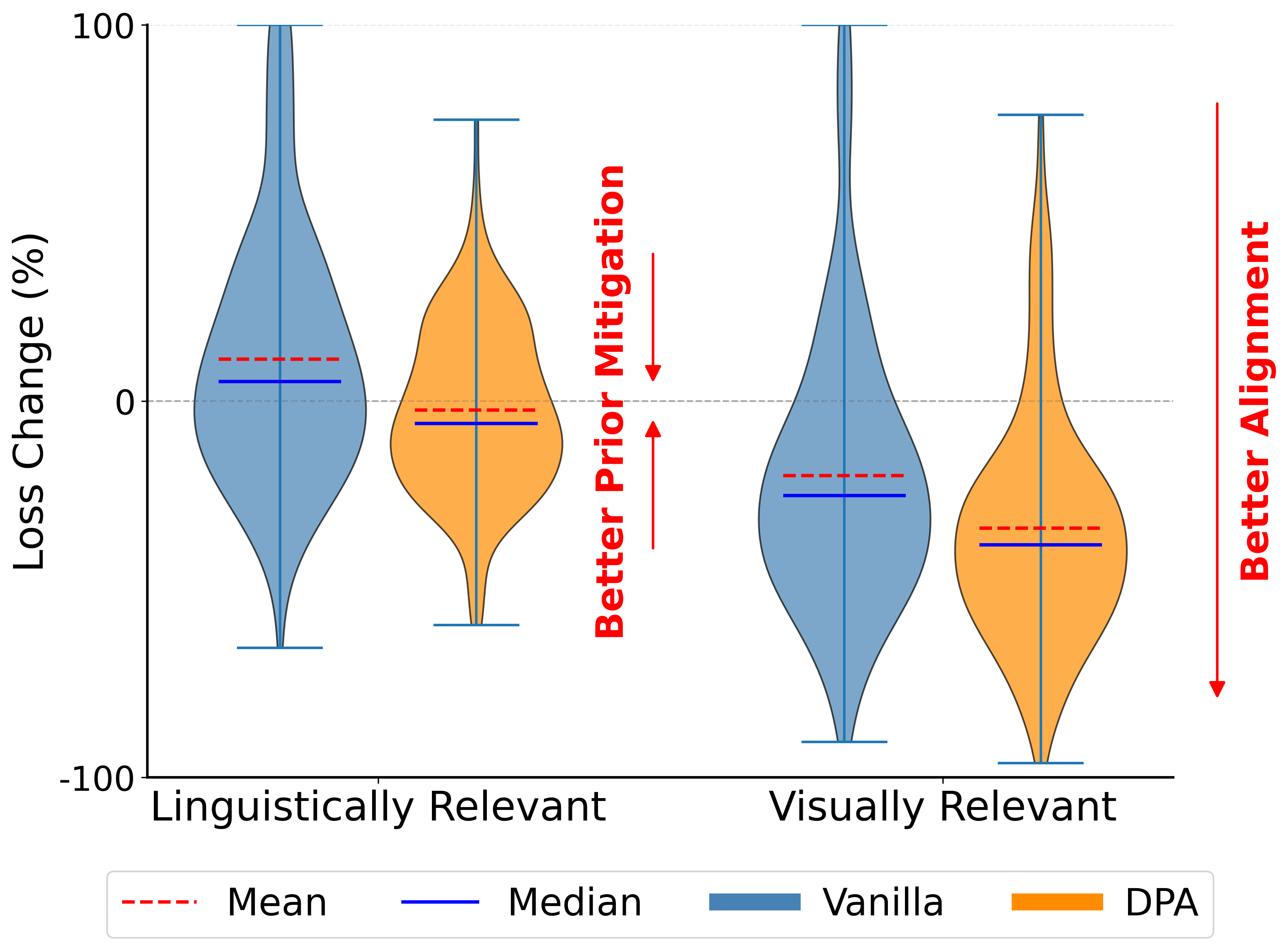}\label{fig:first_2}}
    \caption{In this paper, we identify the issue of language prior conflict. Figure \ref{fig:first_1} illustrates that datasets considered "high-quality" for one model may negatively affect another due to language prior conflict. Figure \ref{fig:first_2} shows that DPA enables models to focus more on vision-text alignment rather than overfitting to language priors in the training data. See Section \ref{sec:main_results} for more analysis.}
\end{figure}


\begin{figure*}[!th]
    \centering
    \includegraphics[width=0.9\linewidth]{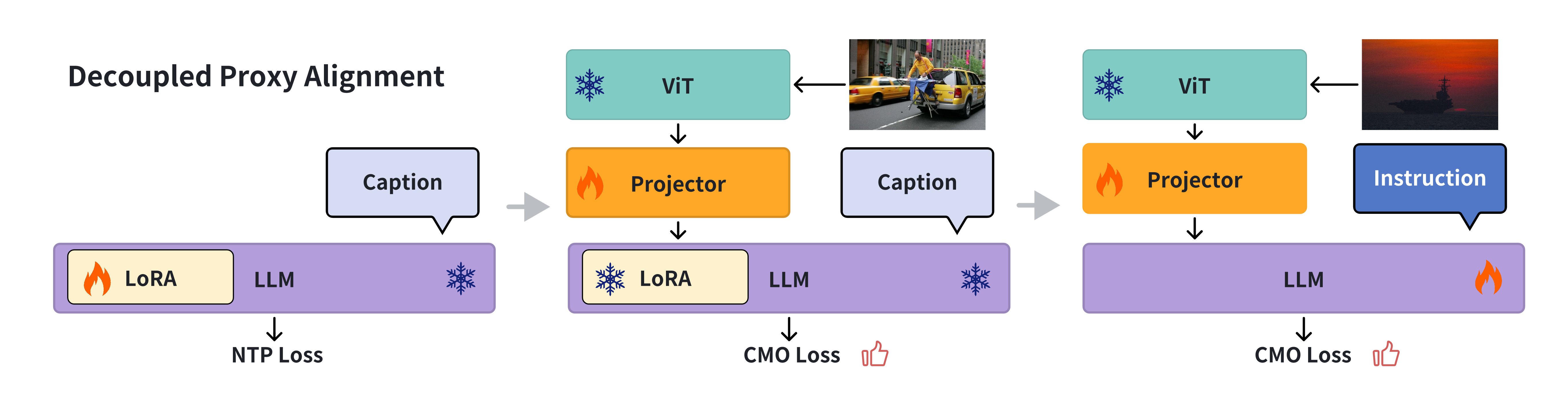}
    \caption{Illustration of Decoupled Proxy Alignment (DPA). From left to right: Proxy LLM Pretraining, Proxy MLLM Pretraining , and MLLM Instruction Tuning. See Section \ref{sec: Methodology-Decoupled Proxy Alignment} for details.}
    \label{fig:method}
    \vspace{-10pt}
\end{figure*}

After the significant success of large language models (LLMs) \cite{dubey2024llama,yang2024qwen2}, numerous efforts have been made to leverage the powerful language understanding capabilities of LLMs to construct multimodal large language models (MLLMs). Many recent studies are centered around enhancing the performance of MLLMs, which can be divided into three categories: (1) introducing high-quality datasets \cite{chen2024sharegpt4v,deitke2024molmo}, (2) improving model architecture design \cite{tong2024cambrian,dubey2024llama}, and (3) optimizing training strategies \cite{xiao2024seeing,chen2024expanding}. These approaches are both distinctive and complementary, collectively driving the development of MLLMs and significantly improving their performance across diverse tasks.


Despite these efforts, our research has uncovered a critical issue during the training of MLLMs: a significant mismatch between the inherent language priors of LLMs and the language priors present in the training datasets. This mismatch causes MLLMs to adapt to the language style of the training samples, which compromises vision-language alignment and results in suboptimal performance. We term this phenomenon as \textbf{language prior conflict}, a challenge that has \textbf{not} been effectively addressed in existing methods. Consequently, there is an urgent need for a more effective multimodal training approach that mitigates the interference of language prior conflict and enhances the alignment between visual and language modalities.

To address this challenge, we propose a novel method called \textbf{Decoupled Proxy Alignment (DPA)}. The core idea of DPA is to decouple the vision-language alignment process from the interference caused by language prior conflicts. Specifically, DPA integrates two key components: (1) During the pretraining phase, a proxy LLM is introduced to mitigate the impact of language prior conflict, ensuring a less biased alignment process. (2) Throughout training, the loss weights are dynamically adjusted to strengthen the optimization signals for visually relevant tokens, rather than those related to linguistic style, further enhancing vision-language alignment.


Experimental results demonstrate that DPA effectively mitigates language prior conflicts and significantly outperforms baseline methods across various model families and training datasets. Furthermore, DPA exhibits strong generalization capabilities, achieving consistently superior performance across datasets and models of varying scales.

Our contributions can be summarized as follows:

\begin{itemize}[itemsep=1pt, leftmargin=10pt, parsep=0pt, topsep=1pt]
\item
We are the first to define and investigate the issue of language prior conflict in MLLMs, experimentally verifying its negative impact on vision-language alignment.
\item
We introduce DPA, a three-stage training method that effectively mitigates language prior conflict and enhances vision-language alignment.
\item
Through extensive experiments, we validate the effectiveness of DPA, demonstrating significant improvements in alignment performance and outstanding generalization capabilities.
\end{itemize}

\section{Related Work}

\paragraph{Multimodal Large Language Models}
Multimodal large language models (MLLMs) have achieved significant advancements in visual understanding, progressing from basic image captioning to complex visual reasoning tasks \cite{li2024llava,Qwen2.5-VL}. These models typically combine a pretrained vision encoder \cite{radford2021learning,zhai2023sigmoid} with a pretrained language model \cite{touvron2023llama,yang2024qwen2}, integrating the two modalities through connectors such as multi-layer perceptrons (MLPs) \cite{liu2024visual,liu2024improved} or cross-attention modules \cite{dai2023instructblip,dubey2024llama}.

A widely adopted training strategy for MLLMs is the two-stage visual instruction tuning framework, first proposed by \citet{liu2024visual}. This methodology has been validated by subsequent studies \cite{agrawal2024pixtral,mckinzie2024mm1,tong2024cambrian}. While recent advancements, such as ShareGPT4V \cite{chen2023sharegpt4v} and InternVL \cite{chen2024expanding}, have introduced more sophisticated training protocols, they are fundamentally built upon the two-stage framework. Given its demonstrated efficacy, our study also adopts this two-stage training method.

\paragraph{Language Prior}

The concept of Language prior refers to the unique linguistic characteristics that LLMs develop during training.
These characteristics include distinct language patterns, styles, vocabularies, grammatical preferences, and implicit world knowledge. Language prior has already attracted significant attention in LLM research. \citet{li2023origin,wang2023seqxgpt} have demonstrated that models with similar language priors exhibit strong behavioral correlations in their predictions, enabling model tracing. Furthermore, \citet{yang2024self,wang2024role} have shown that conflicts in language priors can lead to the forgetting of a model’s original knowledge and capabilities.

In the context of MLLMs, language prior introduces two key challenges: \textbf{(1)} MLLMs are prone to capturing spurious correlations present in multimodal training data \cite{agarwal2020towards,goyal2017making}. \textbf{(2)} 
MLLMs often rely disproportionately on textual prediction, which diminishes their dependence on the visual modality \cite{leng2024mitigating}.
These challenges have significant implications for the performance of multimodal models.


\paragraph{Image-text Modality Alignment} Image-text modality alignment has long been regarded as a core challenge in multimodal understanding. Traditional approaches to image-text alignment often involve training multimodal models from scratch using strategies such as contrastive learning or autoregressive learning \cite{radford2021learning,lin2024rho}. In recent years, 
researchers have made significant strides by leveraging larger and higher-quality datasets, leading to notable advancements in cross-modal alignment \cite{chen2023sharegpt4v,deitke2024molmo}. However, these methods often come at the cost of substantial human and computational resources.
More recently, \citet{xiao2024seeing} proposed CAL, which improves alignment by dynamically adjusting the importance of different tokens during the alignment process, achieving superior results. Despite these improvements, the underlying mechanisms driving these gains remain largely unexplored.
In this study, we present a comprehensive analysis of the conflicts between the language priors in training data and those inherent to LLMs. Furthermore, we propose a novel method designed to effectively mitigate these conflicts.





\section{Language Prior Conflict}
\label{sec: Language Prior Conflic}
In this section, we analyze how language prior conflict impedes the alignment training of multimodal models and ultimately degrades their performance. In Section \ref{sec: Language Prior Conflic-Causes and Consequences}, we start by defining language prior conflict, followed by an exploration of its causes and potential adverse effects. In Section \ref{sec: Language Prior Conflic-Negative Impacts}, we present two quantitative experiments to comprehensively demonstrate the impact of language prior conflict on MLLMs.

\subsection{Causes and Consequences}
\label{sec: Language Prior Conflic-Causes and Consequences}
Language prior conflict refers to the mismatch between the inherent language priors of LLMs and those present in their training datasets. This phenomenon is particularly pronounced in the training of multimodal models. LLMs are typically trained on diverse, large-scale text corpora that cover a wide range of topics and styles. In contrast, image-caption datasets \cite{chen2024sharegpt4v, deitke2024molmo} primarily focus on objective descriptions of visual scenes, often generated by advanced models or through human annotation. These datasets exhibit linguistic distributions that differ significantly from the data used to pretrain LLMs.

During the pretraining phase of MLLMs, the model may prioritize minimizing training loss by adapting to the language style of the training samples rather than focusing on image-text alignment. This prioritization can even lead to severe conflicts, such as overfitting to the style and knowledge contained in the training dataset. 
In the following section, we present experimental evidence demonstrating the impact of language prior conflict on the performance of MLLMs.

\subsection{Negative Impacts}
\label{sec: Language Prior Conflic-Negative Impacts}

\subsubsection{Dataset Quality Paradox}
A surprising discovery in MLLM training is that datasets regarded as "high-quality" for one model may negatively impact another due to conflicts in language priors. We refer to this phenomenon as \textbf{dataset quality paradox}.
To explore this further, we conducted a comparative study using two LLM backbones and two image-caption datasets. More details can be seen in Appendix \ref{sec:appendix_dataset_quality_paradox}.

The experimental results are presented in Figure \ref{fig:first_1}. Vicuna-7B-1.5, trained on the high-quality dataset (ShareGPT4V-PT), demonstrates superior performance compared to the model trained on BLIP-LCS. This aligns with the expectation that high-quality data enhances performance. However, for Qwen2.5-7B-Instruct, training on the high-quality dataset led to a performance decline.

This discrepancy is attributed to a significant conflict between the language priors of Qwen2.5-7B-Instruct and the ShareGPT4V-PT dataset. Specifically, Qwen2.5-7B-Instruct’s advanced language capabilities may cause it to overly focus on textual content in the high-quality dataset, especially when processing lengthy captions, while underutilizing visual information. In contrast, Vicuna-7B-1.5 benefits from its extensive training on open-source GPT-4 distilled data, resulting in linguistic priors that are more compatible with the ShareGPT4V-PT dataset. Additionally, Vicuna’s relatively weaker language capabilities reduce the risk of overfitting on complex captions, encouraging a greater reliance on visual features. This enables Vicuna-7B-1.5 to more effectively learn the correspondence between images and text.

\subsubsection{Quantitative Analysis}
To further investigate the underlying reasons for the performance drop observed when training Qwen2.5-7B-Instruct on ShareGPT4V-PT, we conduct a detailed quantitative analysis of word-level loss changes during training. Specifically, we examine how the model’s loss on linguistically relevant and visually relevant words evolves before and after training on both BLIP-LCS and ShareGPT4V-PT. 
The detailed experimental setup is provided in Appendix \ref{sec:appendix_word_level_loss}.

\begin{figure}[!h]
    \centering
    \includegraphics[width=0.88\columnwidth]{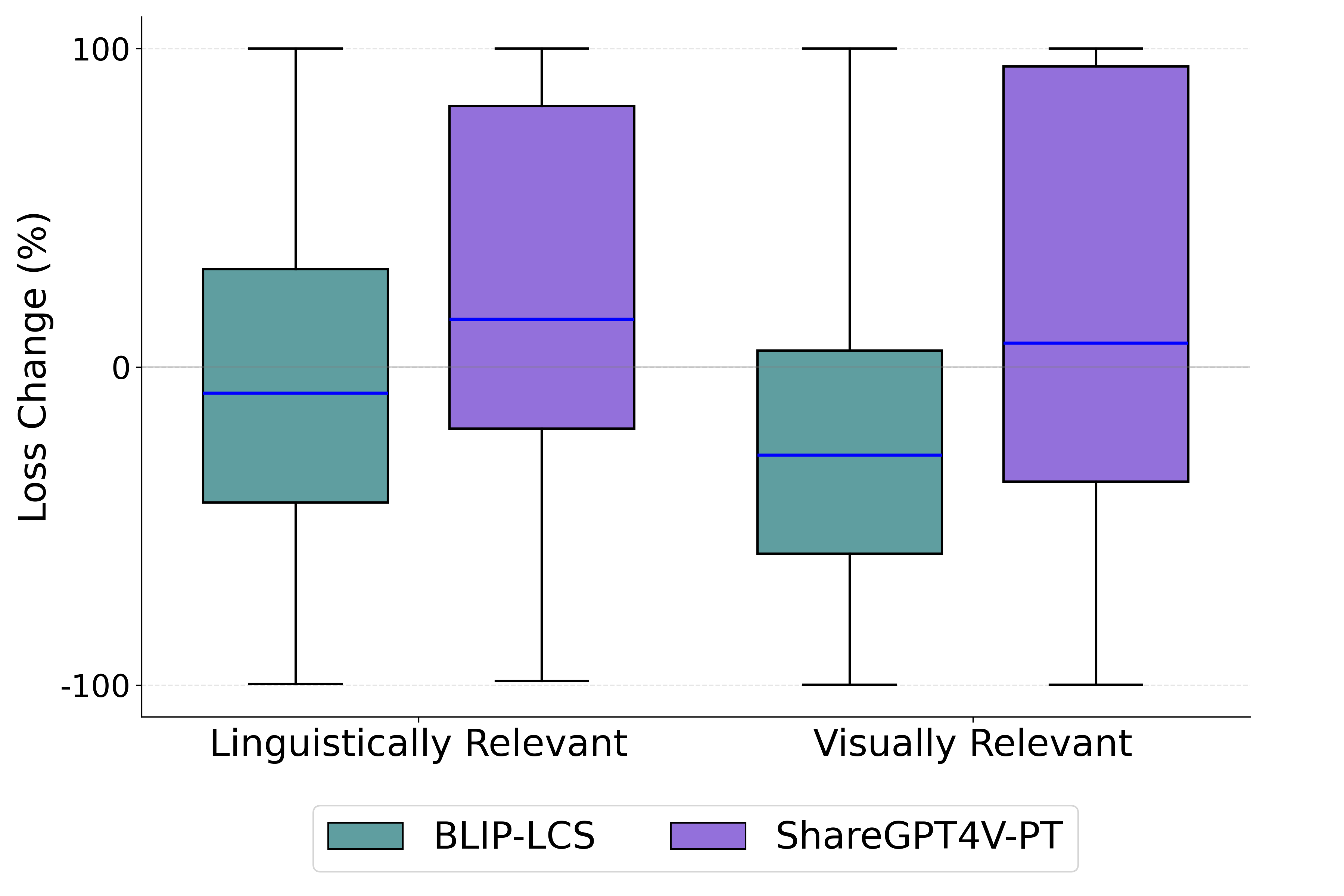}
    \caption{Word-level loss change for linguistically relevant and visually relevant words after training Qwen2.5-7B-Instruct on BLIP-LCS and ShareGPT4V-PT. The results highlight the negative impact of language prior conflict on multimodal alignment.}
    \label{fig:quantitative_loss}
\end{figure}

As shown in the figure \ref{fig:quantitative_loss}, for BLIP-LCS, the loss change of linguistically relevant words centers around zero, indicating that the model does not overfit these words. In contrast, for ShareGPT4V-PT, the loss change for linguistically relevant words fluctuates dramatically due to overfitting of high-frequency words and increased loss for others. For visually relevant words, BLIP-LCS leads to a consistent loss decrease, reflecting effective visual-text alignment, while ShareGPT4V-PT shows an increase in loss, suggesting that language prior conflict hinders multimodal alignment. These results highlight the negative effect of language prior conflict on MLLM training with Qwen2.5-7B-Instruct.

\section{Methodology}
In this section, we detail our three-stage training framework. First, we introduce Proxy Model Optimization (PMO) to mitigate language prior conflicts during pretraining via a proxy LLM. Next, we present Contrastive Modality Optimization (CMO), which enhances visual-language alignment through token reweighting. Finally, we combine these strategies into our \textbf{Decoupled Proxy Alignment (DPA)} method, effectively addressing the key challenges discussed at the end of Section \ref{sec: Language Prior Conflic-Negative Impacts}.

\subsection{Proxy Model Optimization}
\label{sec: Methodology-Proxy Model Optimization}

As described in Section \ref{sec: Language Prior Conflic-Negative Impacts}, during the pretraining phase, when the language priors of LLMs conflict with those of an image-caption dataset, the model tends to focus on textual information while insufficiently utilizing visual information. To address this issue, we propose a novel approach called Proxy Model Optimization (PMO).

The PMO methodology involves a two-stage training process. (1) We train the LLM exclusively on the textual portion of the image-caption dataset. This aligns the LLM’s language priors with the linguistic style and characteristics of the dataset, resulting in a dataset-adapted LLM, referred to as the Proxy LLM. 
(2) We construct a Proxy MLLM by integrating the Proxy LLM, which is kept frozen during subsequent training on the full image-caption dataset. Since the Proxy LLM has already captured the language priors of the training data, the model can focus more effectively on vision-language alignment rather than relearning language priors. This two-stage training process effectively resolves the issue of language prior conflicts during the pretraining phase.


To further optimize the training of the Proxy LLM, we introduce Low-Rank Adaptation (LoRA), a method that offers significant advantages in mitigating catastrophic forgetting and reducing computational overhead. On the one hand, directly training the Proxy LLM on the dataset's textual data may lead to catastrophic forgetting of the original LLM's pretrained knowledge and general language capabilities. LoRA addresses this by freezing the weights of the original LLM and training only a small number of low-rank matrices, enabling the Proxy LLM to adapt to the language priors of the dataset while retaining the core knowledge of the base LLM. On the other hand, compared to full fine-tuning, LoRA saves a significant number of trainable parameters, drastically lowering the computation costs and the training time.

\subsection{Contrastive Modality Optimization}
\label{sec: Methodology-Contrastive Modality Optimization}

To further enhance the alignment between the visual and language modalities during training stage, we propose contrastive modality optimization (CMO). The primary motivation behind CMO lies in the varying degrees of visual relevance among tokens in the captions. For example, compared to nouns, verbs, and adjectives that directly describe visual attributes, function words, discourse markers, and expansive descriptions may contribute less to visual alignment. Treating all tokens equally in loss computation may therefore not be optimal. The design goal of CMO is to dynamically adjust the loss weights to strengthen the optimization signals for visually relevant tokens, rather than those related to linguistic style.

CMO achieves this dynamic token weighting through a contrastive method. During training, for each token in the caption, CMO estimates its visual relevance by comparing the predicted probabilities of the token in two scenarios: (1) when the multi-modal context with the input image is provided and (2) when only the textual context is provided. 
Notably, we directly adjust the weights based on probabilities, which is simpler and more effective compared to previous works such as CAL \cite{xiao2024seeing} that rely on logits. 
Through this contrastive approach, CMO can further decouple the influence of language priors, thus re-evaluating the relevance of the current token to the visual input. Intuitively, tokens with higher visual relevance will exhibit greater differences in predicted probabilities between the scenarios with and without image input. As a result, CMO can effectively capture and amplify the visual alignment signals of these tokens. The detailed algorithm is depicted in Algorithm \ref{code:alt_cmo}.

\newcommand{\ours}{\textit{CMO }}

\begin{center}
\vspace{-0.4cm}
\begin{minipage}{0.46\textwidth}
\begin{algorithm}[H]
\caption{Detail Procedure of $\mathcal{L}_{\ours}$}
\label{code:alt_cmo}
\footnotesize
\renewcommand{\algorithmicrequire}{\textbf{Input:}}
\renewcommand{\algorithmicensure}{\textbf{Output:}}
\begin{algorithmic}[1]
\REQUIRE 
Visual input $V_i$, 
Textual sequence $S = \{s_1, s_2, \ldots, s_m\}$, 
Model distribution $D_\phi$

\STATE Extract probability vectors: \\
\quad$\mathbf{r}^{i,j} = D_{\phi}(V_i,S^{i,<j}), \mathbf{q}^{i,j} = D_{\phi}(S^{i, <j}) $ \\

\STATE Calculate differential score $\delta_s[s_j]^{i,j}$: \\
\quad $\delta_s[s_j]^{i,j} = \mathbf{r}[s_j]^{i,j} - \mathbf{q}[s_j]^{i,j}$ \\

\STATE Transform scores to weights through normalization: \\
\quad $\omega_{i,s_j}^{\prime} = \text{aggregate}_{\Omega}(\text{clip}(\delta_s[s_j]^{i,j}, \alpha, \beta))$ \\
\quad $\omega_{i,s_j} = \frac{\omega_{i,s_j}^{\prime}}{\sum_{k=1}^{m}\omega_{i,s_k}^{\prime}}$ \\

\STATE Formulate final loss through token weighting: \
\quad $\mathcal{L}_{\ours} = -\sum_{i=1}^{N}\sum_{j=1}^{m} \omega_{i,s_j} \log D_{\phi}(s_j|V_i,S^{i,<j})$ \\

\ENSURE Optimized model parameters $\phi^*$

\end{algorithmic}
\label{algo:v2}
\end{algorithm}
\end{minipage}
\end{center}


\subsection{Decoupled Proxy Alignment}
\label{sec: Methodology-Decoupled Proxy Alignment}
The overall methodology comprises three stages:

\begin{itemize}[itemsep=1pt, leftmargin=10pt, parsep=0pt, topsep=1pt]
\item
\textbf{Proxy LLM Pretraining}:
We train the LLM solely on the text portion of the image-caption dataset to obtain a Proxy LLM that is adapted to the language priors of the dataset.
\item
\textbf{Proxy MLLM Pretraining}:
The Proxy LLM is integrated with vision encoder and visual connector layer to construct Proxy MLLM. Then, the complete image-caption dataset is used to train the Proxy MLLM with CMO. In this stage, only the connector layer is trainable.
\item
\textbf{MLLM Instruction Tuning}:
The final MLLM is constructed by combining the original LLM, the pretrained connector layer, and the vision encoder. It then undergoes instruction tuning using CMO, during which both the connector layer and the LLM are trainable.
\end{itemize}
This approach aims to decouple the visual-language alignment process from the potential interference of the language prior conflicts. Additionally, a proxy model is introduced in Stage 1 to enhance the alignment process. Therefore, this method is referred to as \textbf{Decoupled Proxy Alignment}.

\section{Experiments}
In this section, we evaluate the performance of our models through comparative analysis on a variety of visual benchmarks, demonstrating the advantages of our approach. 

Please refer to Appendix~\ref{sec: appendix_detailed_experimental_setup} for detailed experimental setup, including datasets, evaluation metrics, baselines, and implementation details.

\subsection{Main Results}
\label{sec:main_results}

\begin{table*}[!th]
    \centering
    \scalebox{0.8}{
    \begin{tabular}{llcccc c cccc c}
    \toprule
    \multirow{2}*{Dataset} &\multirow{2}*{Method} & \multicolumn{4}{c}{\textbf{General}} & &\multicolumn{4}{c}{\textbf{Vision-centric}} &\multirow{2}*{Avg.}\\
    \cmidrule{3-6} \cmidrule{8-11}
    & &MMB. &OCRB. &DocVQA &AI2D &  &CV-2D &CV-3D &MMVP &NaB. &\\
    \hline
    \multicolumn{12}{c}{\em Qwen2.5-7B-Instruct}\\
    \hline
    \multirow{3}{*}{BLIP-LCS} &Vanilla & 74.5 & 341 & 33.3 & 72.8 & & 59.7 & 66.1 & \textbf{40.0} & 18.8 & 49.9 \\
    &CAL & 74.5 & \textbf{356} & 34.2 & \textbf{74.0} & & 58.4 & 67.1 & 32.0 & 18.4 & 49.3 \\
    &DPA & \textbf{75.8} & 345 & \textbf{34.7} & 72.2 & & \textbf{63.5} & \textbf{68.7} & 38.7 & \textbf{20.5} & \textbf{51.1} \\
    \hline
    \multirow{3}{*}{ShareGPT4V-PT} &Vanilla & 73.1 & 365 & 34.7 & 72.9 & & 58.5 & 62.1 & 36.0 & 18.5 & 49.0 \\
    &CAL & 75.7 & 367 & 35.3 & \textbf{74.7} & & 60.0 & \textbf{70.2} & 38.0 & 17.8 & 51.0 \\
    &DPA & \textbf{76.1} & \textbf{368} & \textbf{36.0} & 74.1 & & \textbf{61.1} & 68.1 & \textbf{40.7} & \textbf{20.4} & \textbf{51.7} \\
    \hline
    \multirow{3}{*}{PixMo-Cap} &Vanilla & 76.4 & 377 & 37.7 & 74.7 & & 61.5 & \textbf{70.4} & 41.3 & 19.8 & 52.4 \\
    &CAL & 76.4 & 392 & 38.0 & \textbf{75.0} & & 61.8 & 69.9 & 42.7 & 19.9 & 52.9 \\
    &DPA & \textbf{77.0} & \textbf{404} & \textbf{38.2} & \textbf{75.0} & & \textbf{67.9} & 65.4 & \textbf{45.3} & \textbf{20.7} & \textbf{53.7} \\
    \hline
    \multicolumn{12}{c}{\em Llama-3.1-8B-Instruct}\\
    \hline
    \multirow{3}{*}{BLIP-LCS} &Vanilla & 70.4 & 319 & 30.1 &\textbf{65.4} & & 58.5 & 59.2 & 28.0 & 13.7 & 44.6 \\
    &CAL & 70.4 & 329 & \textbf{30.6} & 65.1 & & 58.9 & 67.6 & 28.0 & \textbf{16.8} & 46.3 \\
    &DPA & \textbf{70.9} & \textbf{336} & \textbf{30.6} & 65.3 & & \textbf{61.4} & \textbf{68.5} & \textbf{28.7} & 15.4 & \textbf{46.8} \\
    \hline
    \multirow{3}{*}{ShareGPT4V-PT} &Vanilla & 69.4 & \textbf{350} & \textbf{32.3} & 65.8 & & 62.2 & 62.2 & 33.3 & 15.6 & 47.0 \\
    &CAL & 68.3 & 344 & 31.7 & 66.7 & & 57.4 & 65.5 & 32.7 & 10.9 & 46.0 \\
    &DPA & \textbf{71.5} & 346 & 32.2 & \textbf{66.9} & & \textbf{66.1} & \textbf{71.8} & \textbf{39.3} & \textbf{16.6} & \textbf{49.9} \\
    \hline
    \multirow{3}{*}{PixMo-Cap} &Vanilla & 68.4 & 347 & 33.1 & 66.9 & & 60.3 & 60.6 & 37.3 & 16.1 & 47.2 \\
    &CAL  & 71.1 & \textbf{361} & \textbf{34.7} & \textbf{67.4} & & 62.3 & 66.8 & \textbf{38.0} & 17.8 & 49.3 \\
    &DPA & \textbf{72.3} & 349 & 34.4 & 67.2 & & \textbf{63.7} & \textbf{71.4} & 37.3 & \textbf{18.6} & \textbf{50.0} \\
    \hline
    \multicolumn{12}{c}{\em Gemma-2-9B-it}\\
    \hline
    \multirow{3}{*}{BLIP-LCS} &Vanilla & 72.5 & 334 & 31.0 & 67.6 & & 59.4 & 60.9 & \textbf{28.7} & 15.3 & 46.1 \\
    &CAL & 71.9 & \textbf{340} & 31.2 & 67.6 & & 58.3 & \textbf{62.8} & \textbf{28.7} & 15.7 & 46.3 \\
    &DPA & \textbf{72.9} & 335 & \textbf{31.6} &\textbf{ 67.7} & & \textbf{62.7} & 61.5 & 25.3 & \textbf{15.8} & \textbf{46.4} \\
    \hline
    \multirow{3}{*}{ShareGPT4V-PT} &Vanilla & 72.9 & 354 & 34.2 & 68.4 & & 64.0 & 60.5 & 33.3 & 16.7 & 48.2 \\
    &CAL & 71.7 & 362 & 34.3 & 67.6 & & 60.4 & \textbf{61.8} & 30.0 & 17.4 & 47.4 \\
    &DPA & \textbf{74.3} & \textbf{377} & \textbf{35.5} & \textbf{69.6} & &\textbf{ 64.8} & 58.7 & \textbf{38.7} & \textbf{21.1} & \textbf{50.0} \\
    \hline
    \multirow{3}{*}{PixMo-Cap} &Vanilla & 74.3 & 364 & 37.0 & 70.7 & & 64.1 & 65.2 &\textbf{ 45.3} & 19.9 & 51.6 \\
    &CAL & \textbf{75.3} & 367 & 37.4 & \textbf{71.0} & & 65.0 & 66.8 & 34.7 & 19.8 & 50.8 \\
    &DPA & 74.7 & \textbf{383 }& \textbf{37.5} & 70.4 & & \textbf{65.6} & \textbf{70.8} & 42.7 & \textbf{21.4} & \textbf{52.7} \\
    \bottomrule
    \end{tabular}
    }
    \caption{Evaluation results of baselines and DPA. The best performances within each setting are \textbf{bolded}. Abbreviations: MMB.(MMBench), OCRB.(OCRBench), CV-2D(CVBench-2D), CV-3D(CVBench-3D), NaB.(NaturalBench).}
    \label{tab:main_results_1}
\end{table*}

\paragraph{Compared to Vanilla} 
As shown in Table \ref{tab:main_results_1}, after incorporating DPA, MLLMs trained on diverse pretraining data consistently demonstrated significant performance improvements compared to Vanilla, indicating that DPA effectively mitigates the prevalent language prior conflicts existing between different pretraining datasets and different LLMs. \textbf{Notably}, the MLLM trained with DPA on Llama-3.1-8B-Instruct using PixMo-Cap (a high-diversity dataset with multiple expert annotations) achieved an average improvement of 2.8 points compared to Vanilla. This \textbf{highlights} that even with high-quality, well-aligned annotation data, language prior conflicts still exists. By decoupling language priors and modality alignment processes, DPA effectively suppresses the language-dominated overfitting tendency. This fully validates the effectiveness of our method in multimodal alignment.

\paragraph{Compared to CAL}
When compared with the CAL method, DPA also consistently achieves better performance across all mainstream models and datasets. For instance, using the Llama-3.1-8B-Instruct with ShareGPT4V-PT, the DPA method achieves a score of 16.6 on the NaturalBench benchmark, surpassing CAL by 5.7 points. Furthermore, DPA's average performance across metrics exceeds CAL's by 3.9 points.

Notably, CAL exhibits \textbf{inferior} performance compared to even the Vanilla method in several configurations (e.g., Llama-3.1-8B-Instruct with ShareGPT4V-PT). This phenomenon reveals that simply adjusting the loss weight of visually-related tokens is \textbf{insufficient} to decouple language priors from the modality alignment process. Interference from language priors disrupts the optimization trajectory of modality alignment, ultimately leading to performance degradation. These results highlight the unique advantages of DPA in harmonizing language priors with multimodal alignment.

\paragraph{Word-level Loss Analysis}
\label{sec:loss_ana_main}
To further illustrate how DPA alleviates language prior conflict, we conduct a word-level loss analysis by tracking the loss changes for “Linguistically Relevant” and “Visually Relevant” words before and after training with either the Vanilla or DPA method (see Appendix \ref{sec:appendix_word_level_loss_main_results} for details). As illustrated in Figure~\ref{fig:first_2}, DPA substantially reduces loss fluctuations for linguistically relevant tokens compared to Vanilla, indicating a lower tendency to overfit linguistic styles. At the same time, DPA achieves greater loss reductions for visually relevant tokens, signifying improved visual-text alignment. These results demonstrate that DPA effectively re-prioritizes optimization, suppressing language prior overfitting and enhancing multimodal alignment. This also explains why DPA excels even when training on high-quality datasets, overcoming the dataset quality paradox observed with conventional methods.

\paragraph{Conclusion}
In summary, DPA significantly outperforms both Vanilla and CAL in alleviating language prior conflicts and improving overall multimodal performance. The word-level loss analysis further demonstrates that DPA re-prioritizes optimization, effectively suppressing overfitting to linguistic priors while enhancing visual-text alignment. These results validate the generalizability and robustness of DPA, providing a superior and principled solution for multimodal alignment.

\section{Analysis}
\label{sec:analysis}
In this section, we first verify the effectiveness of each component of DPA in multimodal alignment. We then evaluate DPA across different model scales and data sizes. Finally, we analyze the impact of various reweighted loss strategies on multimodal alignment. The detailed experimental setup is provided in Appendix \ref{sec:analysis_setup}.
\subsection{Ablations Studies}
\paragraph{Component Analysis} 
\begin{table}[!t]
\setlength\tabcolsep{1pt}
    \centering
    \scalebox{0.85}{
    \begin{tabular}{lcc}
    \toprule
    Method &General Avg. &Vision. Avg. \\
    \hline
    Vanilla & 54.3 & 43.8 \\
    + PMO & 55.4 &46.1 \\
    + CMO & 55.7 &47.2 \\
    DPA & \textbf{55.8} & \textbf{47.6} \\
    \bottomrule
    \end{tabular}
    }
    \caption{Ablation study on DPA's components.}
    \label{tab:ablation_dpa_sim}
\end{table}
To evaluate the effectiveness of the core components in the DPA framework, we systematically ablated PMO and CMO to train different models. As shown in Table \ref{tab:ablation_dpa_sim}, combining either PMO or CMO with the Vanilla model improves performance on both general benchmarks and vision-centric benchmarks. Notably, CMO achieves greater improvements (from 54.3 to 55.7 on General benchmarks, and from 43.8 to 47.2 on vision-centric benchmarks) compared to PMO (from 54.3 to 55.4 on General benchmarks, and from 43.8 to 46.1 on vision-centric benchmarks), as CMO enhances modality alignment during both pretraining and instruction tuning, whereas PMO only impacts pretraining. Furthermore, combining both PMO and CMO yields additional performance gains. Specifically, as shown in the detailed tables in the appendix \ref{sec: appendix_detailed_results}, DPA significantly outperforms models with only CMO or PMO on benchmarks like MMVP and MMBench.

\paragraph{Stages of conducting CMO within MLLM Training}
CMO can be integrated into both the PreTraining (PT) stage and the Instruction Tuning (IT) stage in existing MLLMs. In this section, we investigate which stage benefits the most from CMO in Table \ref{tab:ablation_cmo_stage}. Our experimental analysis reveals distinct advantages of CMO integration across different training phases: The instruction tuning (IT) stage contributes the majority of performance gains across all evaluated benchmarks, while pretraining (PT) stage integration further enhances model capabilities, particularly demonstrating marked improvements on vision-centric benchmarks.


\noindent
\paragraph{We present more ablation experiments in the Appendix \ref{sec: appendix_more_results}} These include the necessity of LoRA in PMO, the selection of its rank, whether the third stage of DPA continues to use the proxy LLM, as well as ablation studies on the hyper-parameters $[\alpha, \beta]$ in the clamping operation.

\subsection{Results Across Different Model Scales}
\begin{table}[!th]
    \centering
    \scalebox{0.85}{
    \begin{tabular}{llcc}
    \toprule
    Scales &Method &General Avg. &Vision. Avg. \\
    \hline
    \multirow{2}{*}{1.5B} & Vanilla & 48.2 & 36.5 \\
        & DPA & 49.8 & 42.0 \\
    \hline
    \multirow{2}{*}{3B} & Vanilla & 51.5 & 42.4 \\
        & DPA &53.3 &43.3 \\
    \hline
    \multirow{2}{*}{14B} & Vanilla & 55.6 & 46.4 \\
        & DPA & 55.2 & 49.4 \\
    \hline
    \multirow{2}{*}{32B} & Vanilla & 58.2 & 54.6 \\
        & DPA & 58.0 & 56.3 \\
    \bottomrule
    \end{tabular}
    }
    \caption{Performance across different LLM scales.}
    \label{tab:scaling_llm_sim}
\end{table}
\begin{table}[!th]
    \centering
    \scalebox{0.85}{
    \begin{tabular}{llcc}
    \toprule
    Scales &Method &General Avg. &Vision. Avg. \\
    \hline
    \multirow{2}{*}{5\%} & Vanilla & 1.5 & 13.7 \\
        & DPA & 0.4 & 10.0 \\
    \hline
    \multirow{2}{*}{10\%} & Vanilla & 1.6 & 14.3 \\
        & DPA & 0.1 & 13.5\\
    \hline
    \multirow{2}{*}{25\%} & Vanilla & 32.6 & 32.6 \\
        & DPA & 39.5 & 38.0 \\
    \hline
    \multirow{2}{*}{50\%} & Vanilla & 38.9 & 41.8 \\
        & DPA & 42.3 & 44.1 \\
    \hline
    \multirow{2}{*}{100\%} & Vanilla & 39.2 & 46.7\\
        & DPA & 42.5 & 49.1\\
    \bottomrule
    \end{tabular}
    }
    \caption{Performance across different data size.}
    \label{tab:scaling_data_sim}
\end{table}
To verify that DPA is applicable to LLMs of different scales, we further trained MLLMs based on LLMs with various parameter scales and evaluated them on multimodal benchmarks. As shown in Table \ref{tab:scaling_llm_sim}, models of all scales exhibit a stable performance improvement trend on vision-centric benchmarks. For smaller-scale models (e.g., 1.5B), performance improved from 36.5 to 42.0. For larger-scale models (e.g., 14B), performance increased from 46.6 to 49.4. For 32B models, performance rose from 54.6 to 56.3. Considering that unified training hyperparameters were used in the experiments, further adjustments could lead to additional improvements. This phenomenon strongly demonstrates that DPA has impressive adaptability to LLM scales, and its optimization effect is not significantly affected by the number of LLM parameters.

On general benchmarks, performance showed a slight decline as the LLM scale increased. This may be due to the fact that as the parameter count grows, LLMs are more likely to overfit to the linguistic priors from caption data, causing interference from these linguistic priors during inference. In contrast, all the vision-centric tasks are multiple-choice questions, where inference is less affected by such interference, resulting in no performance decline in the metrics.


\subsection{Results Across Different Data Size}
To investigate the impact of data scale on the performance of MLLMs trained with the DPA method, we conducted experiments by adjusting the data volumes for pretraining (PT) and instruction fine-tuning (IT). Specifically, we trained the models using different subsets of the ShareGPT4V-PT and Cambrian-1 datasets.

Table \ref{tab:scaling_data_sim} indicate that data scale is critical to the effectiveness of the DPA method. When the data size is relatively small ($\le$10\%), the performance of the DPA method is lower than that of the baseline model. This is primarily due to insufficient data, which hinders the Proxy LLM from decoupling language priors and limits the MLLM's ability to assess visual relevance. However, when the data volume reaches 25\% or more, the performance of the baseline model improves significantly. The DPA method further enhances modality alignment, leading to additional performance improvements. As the data scale continues to increase, the DPA method provides even greater improvements in modality alignment and overall performance.

In summary, while the DPA method is limited in effectiveness with small data scales, it demonstrates significant advantages with larger data scales, making it highly valuable for improving the performance of multimodal models.



\subsection{Resutls Across Different Reweighted Loss}
To further validate the effectiveness and generalization of our proposed CMO, we conduct a direct comparison with CAL under the same training strategy (PMO) across multiple model backbones. As shown in Table~\ref{tab:dpa_cal}, CMO consistently achieves the best performance on all models, while CAL sometimes even underperforms the baseline. This suggests that CAL’s performance is highly sensitive to the underlying model, likely due to its reliance on model-specific logits distributions. In contrast, CMO demonstrates strong robustness and generalization, benefiting from its probability-based design. These results highlight the practical advantage of CMO for multimodal model training across diverse architectures.
\begin{table}[!th]
    \centering
    \scalebox{0.8}{
    \begin{tabular}{lccc}
    \toprule
    Method & General. Avg. &Vision. Avg. &Avg.\\

    \hline
    \multicolumn{4}{c}{\em Qwen2.5-7B-Instruct}\\
    \hline
    PMO & 55.4 & 	46.0&	50.7 \\
    PMO + CAL & \textbf{56.0} & 46.9 & 51.5  \\
    PMO + CMO (Ours) & 55.8 &	\textbf{47.6}&	\textbf{51.7}  \\
    \hline
    \multicolumn{4}{c}{\em Llama-3.1-8B-Instruct}\\
    \hline
    PMO &49.2& 	42.4 &	45.8 \\
    PMO + CAL &50.3 &44.1 &	47.2  \\
    PMO + CMO (Ours) & \textbf{51.3} &\textbf{48.5} &\textbf{49.9}  \\

    \hline
    \multicolumn{4}{c}{\em Gemma-2-9B-it}\\
    \hline
    PMO &53.7& 	44.7& 	49.2 \\
    PMO + CAL & 53.5 &	44.3 &	48.9   \\
    PMO + CMO (Ours) & \textbf{54.3} &	\textbf{45.8} &	\textbf{50.0}  \\

    \bottomrule
    \end{tabular}
    }
    \caption{Comparison between our proposed CMO loss and CAL loss when combined with PMO. All models are trained on ShareGPT4V-PT dataset. The best performances within each setting are \textbf{bolded}.}
    \label{tab:dpa_cal}
\end{table}

\section{Conclusion}

In this paper, we introduced the concept of language prior conflict and proposed a novel method, Decoupled Proxy Alignment (DPA), to effectively address this challenge and enhance the alignment between visual and language modalities. Extensive experiments demonstrate that DPA significantly reduces the negative impact of language prior conflict, achieving superior alignment performance across a wide range of datasets, model families, and scales. It not only enhances the training efficiency of MLLMs but also shows exceptional generalization capabilities, making it a robust approach for vision-language alignment.

\section*{Limitations}
While our proposed DPA demonstrates significant improvements in mitigating language prior conflicts and enhancing vision-language alignment, certain limitations remain. Specifically, the selection of lower and upper bounds in CMO process is currently determined empirically, which could be extended to more adaptive settings in further explorations.

\section*{Acknowledge}
This work was supported by the National Natural Science Foundation of China (No. U24B20181) and Shanghai Pilot Program for Basic Research - Fudan University 21TQ1400100 (22TQ018).


\bibliography{custom}

\clearpage
\appendix

\section{Experimental Details}

\subsection{Dataset Quality Paradox}
\label{sec:appendix_dataset_quality_paradox}
To explore the dataset quality paradox, we conducted a comparative study using two LLM backbones: 
\begin{itemize}[itemsep=1pt, leftmargin=10pt, parsep=0pt, topsep=1pt]
\item
\textbf{Vicuna-7B-1.5} A relatively weaker model in text generation.
\item
\textbf{Qwen2.5-7B-Instruct} A model with strong text generation capabilities.
\end{itemize}
and two image-caption datasets:
\begin{itemize}[itemsep=1pt, leftmargin=10pt, parsep=0pt, topsep=1pt]
\item
\textbf{BLIP-LCS} A noisier dataset with shorter captions, commonly used in LLaVA-1.5 pretraining.
\item
\textbf{ShareGPT4V-PT} A high-quality dataset featuring longer, more detailed captions generated by GPT-4.
\end{itemize}

Both models were fine-tuned based on the LLaVA-1.5 architecture under consistent experimental settings and hyperparameters \cite{liu2024visual}. The Cambrian-1 dataset was utilized as the instruction-tuning dataset. Performance evaluation was conducted using CVBench, a vision-centric benchmark specifically designed to account for sensitivity to language priors.

\subsection{Analysis of Word-Level Loss}
\label{sec:appendix_word_level_loss}

To further investigate the impact of language prior conflict on MLLM training, we conducted a word-level loss analysis based on the fine-tuning experiments of Qwen2.5-7B-Instruct on BLIP-LCS and ShareGPT4V-PT.

Specifically, we randomly sampled 100 examples from each dataset. Each example was tokenized at the word level, and GPT-4.1 was used to classify each word as either \textit{Language Prior} or \textit{Visually Relevant}. For each word, we computed its loss before and after fine-tuning. If a word was split into multiple tokens, we used the loss of the first token as the word-level loss. The percentage change in loss for each category was then calculated to analyze the model’s tendency to fit language priors versus visually grounded content.

All other training settings were kept consistent with the main experiments described above.

\subsection{Analysis of Word-Level Loss for Main Results}
\label{sec:appendix_word_level_loss_main_results}

The experimental setup for the word-level loss analysis in Section~\ref{sec:loss_ana_main} closely follows the procedure described in Section~\ref{sec:appendix_word_level_loss}. Specifically, this analysis is based on the fine-tuning results of Qwen2.5-7B-Instruct on the BLIP-LCS dataset.

The only difference from Section~\ref{sec:appendix_word_level_loss} is that, to facilitate clearer visualization, we excluded words with a frequency less than 3 in the sampled data.

All other experimental settings remain consistent with those described above.

\subsection{Ablation Study and Analysis}
\label{sec:analysis_setup}
Unless otherwise specified, all experiments in Section \ref{sec:analysis} are conducted using the Qwen2.5-Instruct series models trained on the BLIP-LCS dataset. If the model size is not explicitly mentioned, Qwen2.5-7B-Instruct is used by default.

\subsection{Detailed Experimental Setup}
\label{sec: appendix_detailed_experimental_setup}

\paragraph{Datasets}
During the pretraining stage, we selected three representative multi-modal pretraining datasets for comparative analysis:
\textbf{BLIP-LCS} \footnote{\textit{LCS} abbreviates the LAION, CC, and SBU datasets}\cite{li2022blip}: Used as the pretraining dataset for LLaVA-1.5\cite{liu2024improved}. It is a filtered subset of LAION\cite{schuhmann2021laion}, CC\cite{sharma2018conceptual}, and SBU\cite{saleh2015large}, with a more balanced distribution of concept coverage. 
\textbf{ShareGPT4V-PT}\cite{chen2024sharegpt4v}: Used as the pretraining dataset for ShareGPT4V\cite{chen2024sharegpt4v}. It utilizes high-quality image-text descriptions generated by the GPT-4, offering significantly richer semantics and contextual coherence compared to BLIP-LCS.
\textbf{PixMo-Cap}\cite{deitke2024molmo}: Used as the pretraining dataset for Molmo\cite{deitke2024molmo}, a state-of-the-art open-source MLLM. This dataset is constructed by expert annotations, featuring precise visual attribute labeling and complex scene descriptions.

During the instruction tuning stage, we follow \citet{tong2024cambrian} and use the Cambrian-1\footnote{\url{https://huggingface.co/datasets/nyu-visionx/Cambrian-10M/blob/main/jsons/Cambrian737k.jsonl}} dataset as our training data. This dataset builds upon LLaVA-665k\cite{liu2024improved}, systematically expanding the model's understanding of structured visual information by incorporating OCR data and chart data.

\paragraph{Evaluation metrics}
We employs a dual-dimensional evaluation system: \textbf{General benchmarks} and \textbf{Vision-centric benchmarks}. General benchmarks include MMBench\cite{liu2024mmbench} (commonsense reasoning), AI2D\cite{hiippala2021ai2d} (diagram parsing), DocVQA\cite{hudson2019gqa} (document understanding) and OCRBench\cite{liu2023hidden} (OCR capability), covering the assessment of fundamental cognitive abilities. Vision-centric benchmarks focus on evaluating core visual capabilities, comprising three specialized test sets: CVBench\cite{tong2024cambrian} examines structured visual understanding through 2D/3D spatial relationship analysis, MMVP\cite{tong2024eyes} emphasizes fine-grained feature recognition, and NaturalBench\cite{li2024naturalbench} tests comprehensive visual perception capabilities through challenging tasks such as understanding attribute bindings and reasoning about object relationships.


We use VLMEvalKit\cite{duan2024vlmevalkit} for systematic evaluation. Specifically, multiple choice questions (AI2D / MMBench / CVBench / MMVP) primarily use the accuracy of the options as the core metric. Document parsing (DocVQA) uses Normalized Levenshtein Distance. OCR recognition (OCRBench) is based on the hit rate of detections contained in the ground-truth answers. Cross-combination evaluation (NaturalBench) sets four fine-grained metrics, including single-question accuracy, group accuracy (requiring all four combined questions to be correct), Question Accuracy (correct rate for the same question on both images) and image accuracy (correct rate for the same image on both questions).

\paragraph{Baselines}
We compare DPA with two representative training paradigms: 
\textbf{(1) Vanilla} is the classic two-stage alignment method \cite{liu2024visual}, where only the MLP projection layer is unfrozen during pretraining, while both the MLP and LLM are unfrozen during fine-tuning, 
\textbf{(2) CAL}\cite{xiao2024seeing} introduces a dynamic weight adjustment mechanism on top of Vanilla, optimizing the loss weights of different tokens through logits differences to enhance alignment of key semantics.

\paragraph{Implementation Details}
The model architecture employs CLIP-pretrained ViT-L as the visual encoder, a two-layer MLP cross-modal connection layer with GeLU activation, and three different LLMs: Qwen2.5-7B-Instruct\cite{yang2024qwen2}, Llama-3.1-8B-Instruct\cite{dubey2024llama}, and Gemma-2-9B-it\cite{team2024gemma}.  The training parameters were optimized through grid search, with learning rates set to 2e-3 and 4e-5 for the pretraining and fine-tuning stages, respectively. The learning rate for the Proxy LLM pretraining phase was set to 4e-5. The batch size was fixed at 256. The LoRA configuration uses rank=256, alpha=512, and weight boundaries $\alpha$=0.05, $\beta$=0.5, with a pooling layer window size of 3. Experiments were conducted on 8 NVIDIA H100 GPUs.

\section{More Experimental Results}

\subsection{Ablation Studies}
\label{sec: appendix_more_results}

\paragraph{Necessity of LoRA}
\begin{figure}
    \centering
    \includegraphics[width=0.85\linewidth]{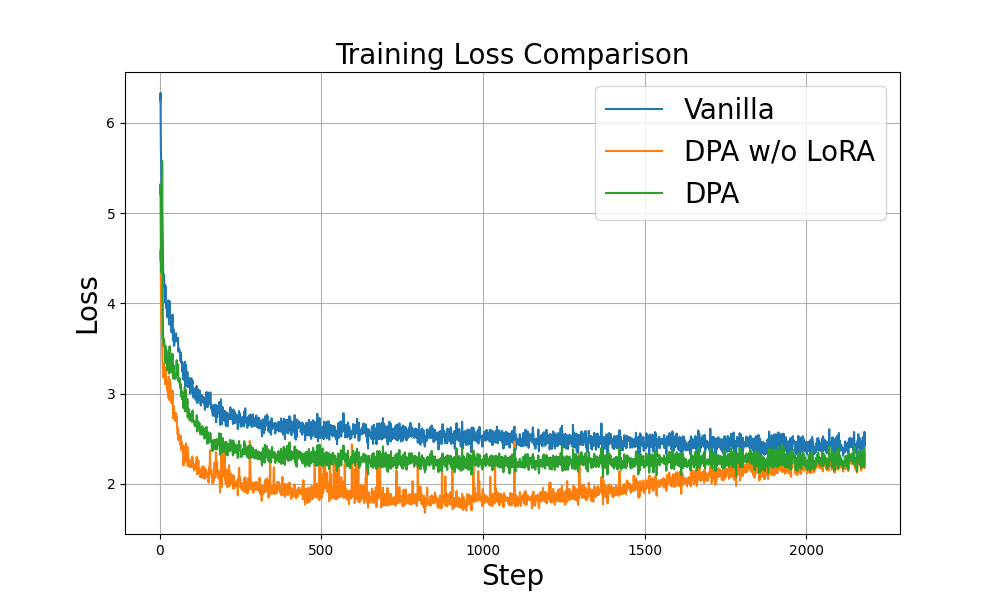}
    \caption{Comparison of pretraining loss function behavior across three training strategies: Vanilla, DPA without LoRA, and DPA with LoRA. }
    \label{fig:loss}
    \vspace{-10pt}
\end{figure}
To systematically validate the effectiveness of the LoRA-enhanced training strategy, this study investigates the pretraining loss function behavior of three training strategies: Vanilla, DPA w/o LoRA, and DPA (with LoRA). As shown in Figure \ref{fig:loss}, the DPA w/o LoRA strategy exhibits a clear non-monotonic convergence pattern: the training loss initially decreases rapidly but then unexpectedly increases, a significant deviation from the typical training curve. This phenomenon indicates that DPA w/o LoRA overfits to a subset of text descriptions during the Proxy Model Optimization phase—the model's loss decreases abnormally on specific samples while its generalization performance degrades significantly on others.

This overfitting phenomenon has a dual negative impact. First, the LLM focuses excessively on local features within the text data rather than learning the overall prior language distribution. Second, in the subsequent visual modality alignment phase, the model struggles to disentangle the already solidified text feature representations, leading to semantic mismatches during cross-modal alignment. Notably, the introduction of LoRA results in the expected monotonic convergence characteristic, validating its effectiveness in suppressing overfitting. 

\paragraph{Rank for LoRA}
\begin{table}[!th]
    \centering
    \scalebox{0.72}{
    \begin{tabular}{lccc}
    \toprule
    Method &LoRA rank &General Avg. &Vision. Avg. \\
    \hline
    DPA w/o LoRA & -  & \textbf{54.2} & 44.5\\
    \hline
    \multirow{3}{*}{DPA with LoRA} &128 &53.3 & 43.5 \\
    &256 &\textbf{54.2} & \textbf{45.7} \\
    &512 &53.8 &44.4 \\
    \bottomrule
    \end{tabular}
    }
    \caption{Performance difference when applying different rank for LoRA. The LoRA alpha is set to twice the LoRA rank.}
    \label{tab:ablation_rank_sim}
\end{table}
Table \ref{tab:ablation_rank_sim} using the Qwen2.5-7B-Instruct model on the BLIP-LCS dataset demonstrates a non-linear relationship between LoRA rank and model performance. As rank increases, metrics on both general and vision-centric benchmarks initially improve, then decline. This phenomenon can be explained as follows: Initially, increasing the rank appropriately increases the number of trainable parameters, enhancing the language model's ability to fit the textual descriptions. This, in turn, allows the model to focus more on semantic matching with the visual modality during cross-modal alignment (DPA). However, when the rank exceeds a certain threshold, the excessive degrees of freedom lead to the model overfitting the textual descriptions, ultimately weakening the visual-language modality alignment.

\paragraph{Proxy LLM vs. Target LLM in Instruction Fine-tuning}
\begin{table}[!th]
    \centering
    \scalebox{0.8}{
    \begin{tabular}{lcccc}
    \toprule
    Initialization &General Avg. &Vision. Avg. & Avg. \\
    \hline

    Proxy LLM& 54.7 & 46.8 & 50.7  \\
    Target LLM & 54.3 & 47.9 & 51.1 \\
    \bottomrule
    \end{tabular}
    }
    \caption{Performance comparison between models initialized with the proxy LLM and the target LLM for instruction fine-tuning.}
    \label{tab:ablation_replace_sim}
\end{table}
To investigate the effect of restoring the proxy LLM to the target LLM in stage 3, we conducted comparative experiments analyzing the performance differences when using different LLMs as starting points for the instruction-tuning phase. Table \ref{tab:ablation_replace_sim} shows that using the target LLM as the starting point significantly improves the model performance on vision-centric benchmarks, while there is a slight decrease on general benchmarks. The experimental results indicate that restoring the proxy LLM to the target LLM in stage 3 is more beneficial for vision-text modality alignment.

\begin{figure*}[!h]
        \centering
        \subfloat[Qwen2.5-7B-Instruct]{\includegraphics[width=.98\columnwidth]{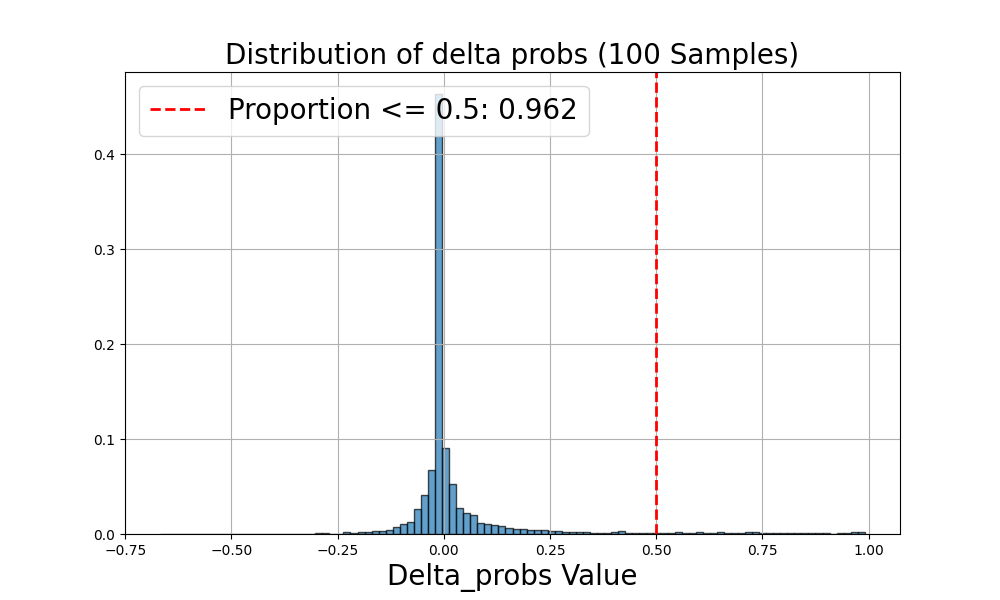}}\hspace{4pt}
        \subfloat[Llama-3.1-8B-Instruct]{\includegraphics[width=.98\columnwidth]{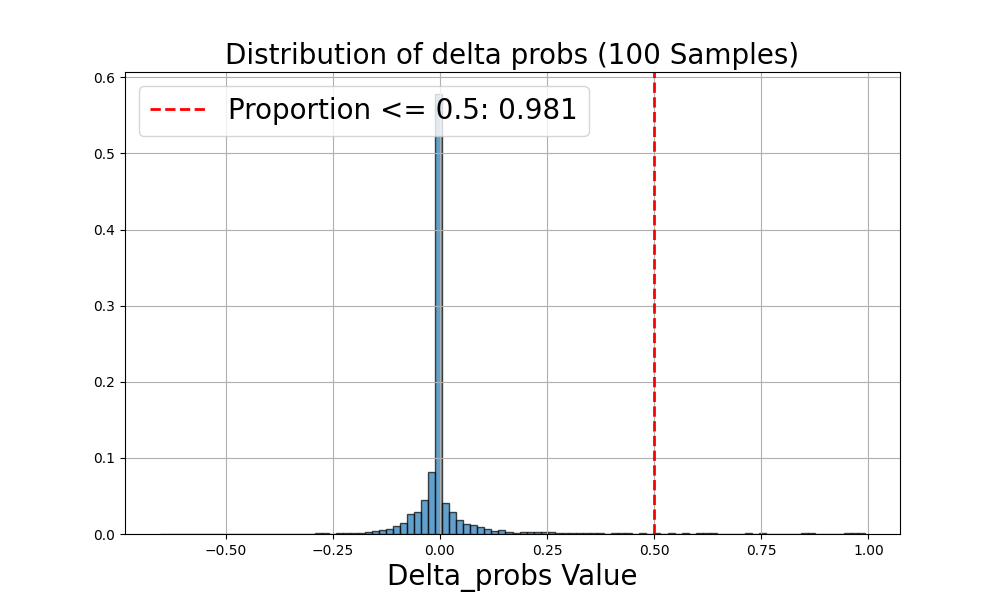}}\hspace{4pt}
        \caption{$\Delta \mathbf{p}$ distribution for models on 100 random sampled cases.}
        \label{figure:weight_bar}
\end{figure*}

\paragraph{Hyper-parameters for $[\alpha,\beta]$ in clamping}
We further conduct an ablation study on $\alpha$ and $\beta$ to study the effect of the hyperparameters in our clamping operation. First, we plot the $\delta$ distribution on MLLMs in Figure \ref{figure:weight_bar}. Tokens whose $\delta$ lower than 0.5 constitute approximately 96\% of the total label sequences.  Based on this observation, we set 0.5 as the upper bound $\beta$. To prevent language style-related tokens from being completely ignored, we set the lower bound $\alpha$ to 0.05. This is because a $\delta$ of 0 implies that the weights of these tokens are zero, meaning they will not be optimized. We then extended both the lower and upper bounds to their extreme values, i.e., 0 and 1. The results are presented in Table \ref{tab:ablation_cmo_hyp}. 
(1) When the upper bound $\beta$ is set to 1, the model's performance degrades significantly. This indicates that allowing a few visually correlated tokens to dominate the importance weights across all label tokens negatively impacts the model. A possible explanation is that these tokens become over-optimized, while other tokens are ignored.
(2) When the lower bound $\alpha$ is set to 0, the model also shows a performance drop. This suggests that focusing solely on optimizing visually correlated tokens is harmful. Instead, the optimization process should cover all tokens while emphasizing visually correlated tokens.

\subsection{Dynamic Resolution Input}
\begin{figure}[t]
    \centering
    \includegraphics[width=.8\linewidth]{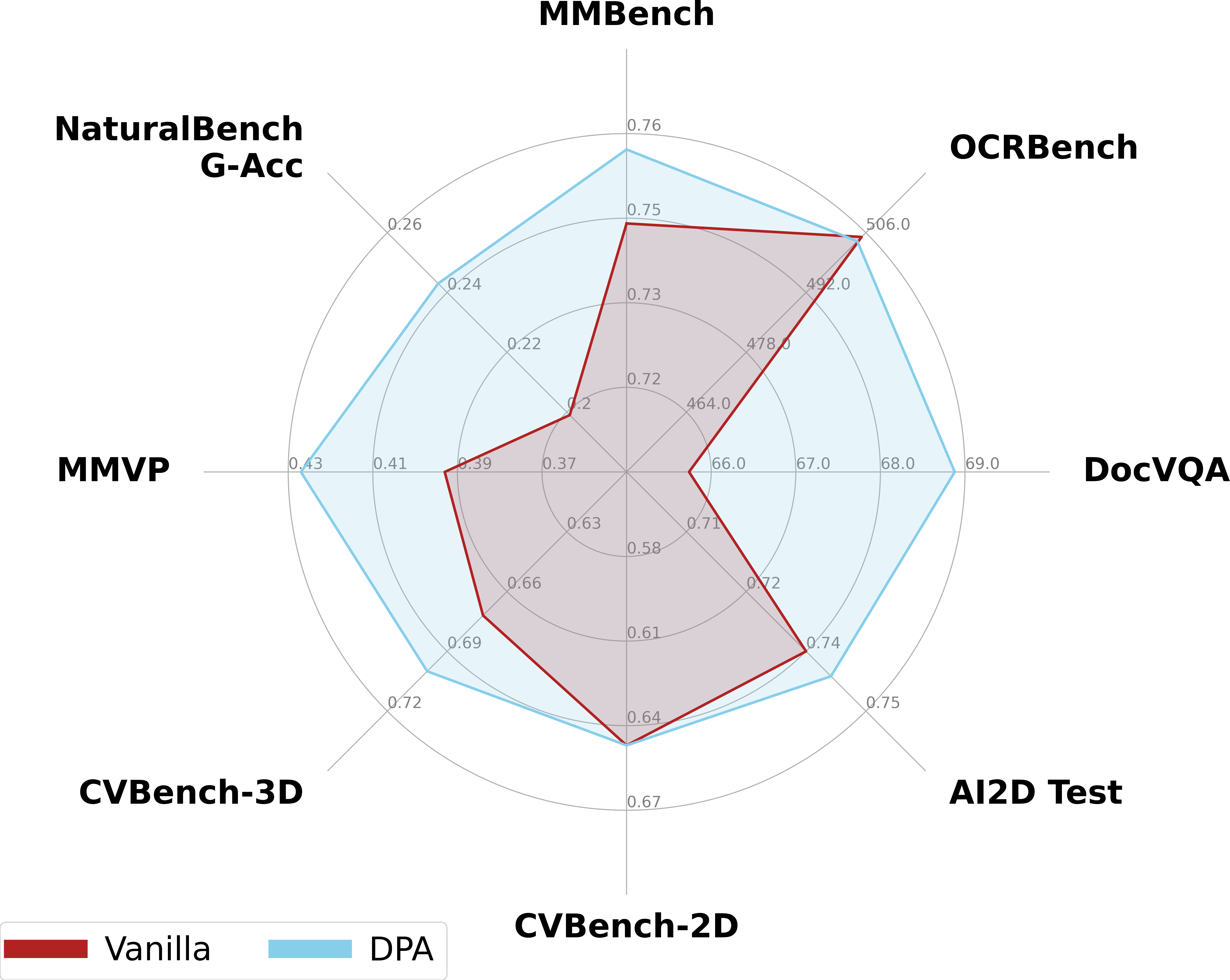}
    \caption{Anyres performance when using DPA or not.}
    \label{fig:anyres}
\end{figure}

Supporting dynamic resolution input is a trend in MLLMs. Based on the Qwen2.5-7B-Instruct model architecture, we experimented with the Anyres training strategy of LLaVA 1.5 on the BLIP-LCS dataset. As shown in Figure \ref{fig:anyres}, the experimental results demonstrate that this method can effectively accommodate dynamic resolution input schemes.

\subsection{Training on Multi-Datasets}
\begin{table*}[!th]
    \centering
    \scalebox{0.85}{
    \begin{tabular}{lcccccccccc}
    \toprule
    Method & MMB. & OCRB. & DocVQA & AI2D & CV-2D & CV-3D & MMVP & NaB. & Avg. \\
    \midrule
    Vanilla & 0.754 & 372 & 36.898 & 0.738 & 0.597 & 0.687 & 0.327 & 0.188 & 51.52 \\
    DPA     & 0.755 & 394 & 37.229 & 0.736 & 0.619 & 0.695 & 0.453 & 0.210 & 54.34 \\
    \bottomrule
    \end{tabular}
    }
    \caption{Performance comparison of Vanilla and DPA on a mixed dataset (BLIP-LCS, ShareGPT4V-PT, and PixMo-Cap, 600K samples total). DPA demonstrates superior average performance and significant gains in vision-centric tasks, indicating effective adaptation to multiple conflicting language priors.}
    \label{tab:multi_prior}
\end{table*}
To address the challenge of training on multiple datasets with conflicting language priors, we conducted an additional experiment by mixing 200K samples each from BLIP-LCS (web-crawled), ShareGPT4V-PT (GPT-generated), and PixMo-Cap (expert-annotated) into a composite dataset containing 600K samples. This mixed dataset naturally introduces diverse and potentially conflicting language priors. We trained Qwen2.5-7B-Instruct on this dataset and compared the performance of DPA against the Vanilla baseline. As shown in Table~\ref{tab:multi_prior}, DPA outperformed Vanilla by a notable margin (average score: 54.34 vs. 51.52), with especially significant improvements in vision-centric tasks such as MMVP (0.453 vs. 0.327). These results indicate that proxy LLM pretraining with LoRA can effectively adapt to diverse language priors while preserving the base LLM’s knowledge, thereby mitigating interference and reducing the risk of catastrophic forgetting.

\subsection{Computational Overhead}
Our proposed PMO requires an additional round of pre-training on the dataset in conjunction with LoRA. For CMO, each iteration involves two forward passes of text tokens. Table \ref{tab:com_overhead} presents the training time of Llama-3.1-8B-Instruct on BLIP-LCS using 8 H100 GPUs. DPA introduces approximately 33\% additional training time. Memory usage increased by approximately 20\% due to LoRA and CMO computations. Given the performance improvements achieved by DPA (e.g., +2.8 on PixMo-Cap, as shown in Table \ref{tab:main_results_1}), this trade-off is considered justified.
\begin{table}[!h]
    \centering
    \scalebox{0.78}{
    \begin{tabular}{lccccc}
    \toprule
    Method &Other&Pretraining &Instruction Tuning & Overall \\
    \hline
    Vanilla & - & 1.37h& 6h & 6.37h\\
    DPA & 0.5h & 1.85h & 6.12h & 8.47h\\

    \bottomrule
    \end{tabular}
    }
    \caption{Training time of different methods (Vanilla and DPA) for pretraining and instruction tuning of Llama-3.1-8B-Instruct on BLIP-LCS using 8 H100 GPUs.}
    \label{tab:com_overhead}
\end{table}

\subsection{Qualitative Analysis of Word-level loss change}
\begin{figure*}[!h]
    \centering
    \includegraphics[width=.8\linewidth]{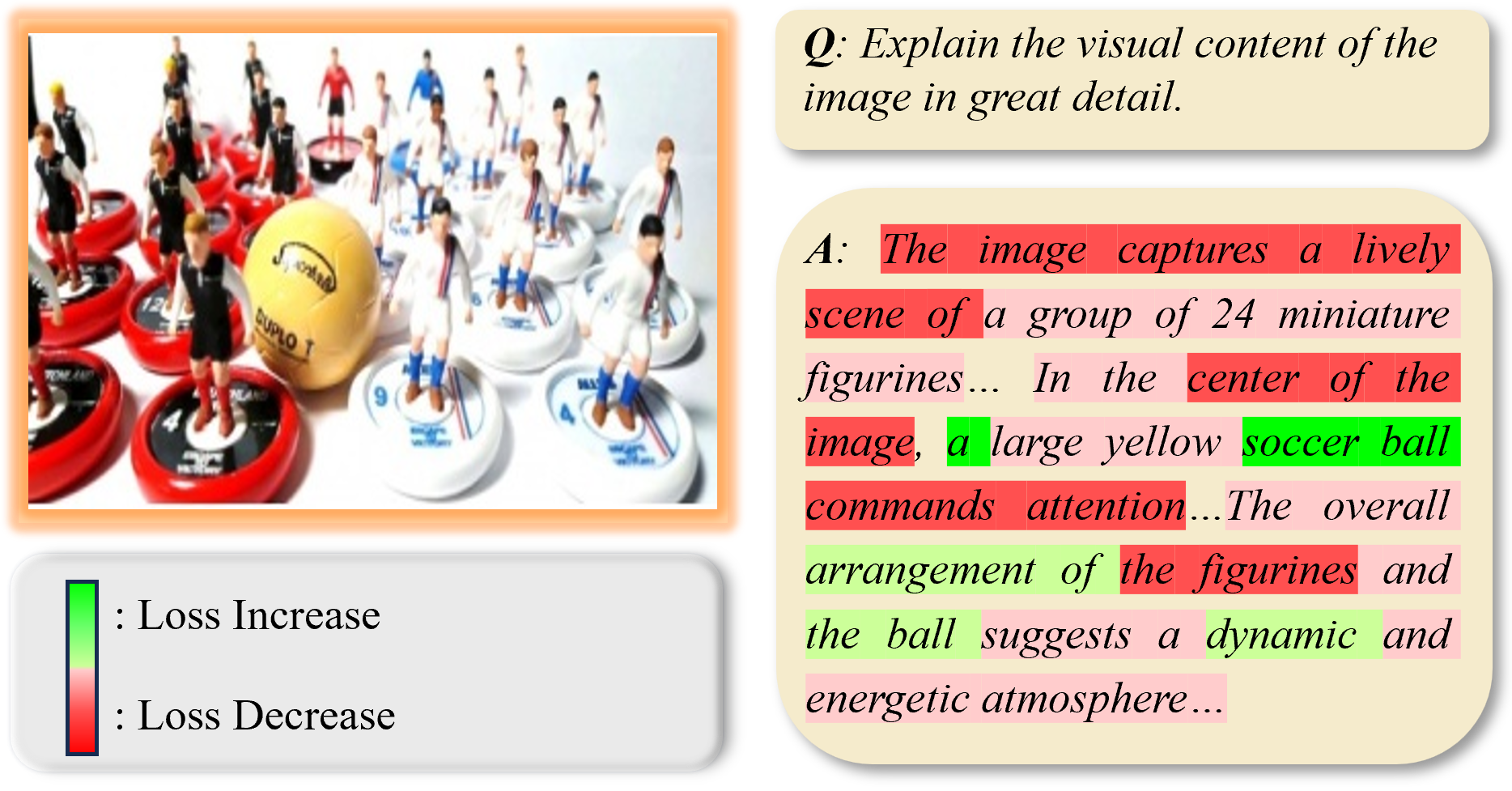}
    \caption{Qualitative Analysis of Word-level loss change.}
    \label{fig:qualitative_word_loss}
\end{figure*}

To further illustrate the impact of language prior conflict, we present a qualitative analysis of word-level loss changes based on a sample from the ShareGPT4V-PT dataset, as shown in Figure~\ref{fig:qualitative_word_loss}. In the figure, words highlighted in red indicate a decrease in loss after fine-tuning, while those in green indicate an increase in loss.

It can be observed that many words with decreased loss are primarily related to language style, such as “captures” and “a lively scene of.” In contrast, some words with increased loss are highly relevant to visual content, such as “soccer” and “ball.” This case further supports our quantitative findings: language prior conflict leads to suboptimal performance when training Qwen2.5-7B-Instruct with ShareGPT4V-PT, as the model tends to fit language priors at the expense of visually grounded content.

\subsection{Detailed Results}
\label{sec: appendix_detailed_results}
\begin{table*}[!th]
    \centering
    \scalebox{0.8}{
    \begin{tabular}{lcccc c cccc}
    \toprule
    \multirow{2}*{Method} & \multicolumn{4}{c}{\textbf{General}} & &\multicolumn{4}{c}{\textbf{Vision-centric}}\\
    \cmidrule{2-5} \cmidrule{7-10}
    &MMB. &OCRB. &DocVQA &AI2D &  &CV-2D &CV-3D &MMVP &NaB.\\
    \hline

    Vanilla & 73.1 & 365 & 34.7 & 72.9 & & 58.5 & 62.1 & 36.0 & 18.5 \\
    DPA w/o PMO & 75.2 & 385 & 35.9 & 73.1 & & 60.6 & 69.1 & 39.3 & 19.8 \\
    DPA w/o CMO & 74.9 & 380 & 35.2 & 73.4 & & 59.1 & 65.3 & 42.0 & 17.8 \\
    DPA & 76.1 & 368 & 36.0 & 74.1 & & 61.1 & 68.1 & 40.7 & 20.4 \\
    \bottomrule
    \end{tabular}
    }
    \caption{Ablation study on DPA's components.}
    \label{tab:ablation_dpa}
\end{table*}
\begin{table*}[!th]
    \centering
    \scalebox{0.8}{
    \begin{tabular}{cc|cccc c cccc}
    \toprule
    \multirow{2}*{PT} &\multirow{2}{*}{IT} & \multicolumn{4}{c}{\textbf{General}} & &\multicolumn{4}{c}{\textbf{Vision-centric}}\\
    \cmidrule{3-6} \cmidrule{8-11}
    &&MMB. &OCRB. &DocVQA &AI2D &  &CV-2D &CV-3D &MMVP &NaB.\\
    \hline
    &&72.4 & 352 & 33.9 & 73.7 & & 60.1 & 67.5 & 37.3 & 17.4 \\
    $\checkmark$ & & 73.2 & 357 & 34.1 & 72.8 & & 57.5 & 67.7 & 40.7 & 18.5 \\
    & $\checkmark$ & 74.5 & 368 & 35.0 & 72.9 & & 61.5 & 69.2 & 39.3 & 18.6 \\
    $\checkmark$ & $\checkmark$ & 75.8 & 345 & 34.7 & 72.2 & & 63.5 & 68.7 & 38.7 & 20.5\\
    \bottomrule
    \end{tabular}
    }
    \caption{Performance difference when \ours is applied at different training stages.}
    \label{tab:ablation_cmo_stage}
\end{table*}
\begin{table*}[!th]
    \centering
    \scalebox{0.8}{
    \begin{tabular}{lcccc c cccc}
    \toprule
    \multirow{2}{*}{$[\alpha, \beta]$} & \multicolumn{4}{c}{\textbf{General}} & &\multicolumn{4}{c}{\textbf{Vision-centric}}\\
    \cmidrule{2-5} \cmidrule{7-10}
    &MMB. &OCRB. &DocVQA &AI2D &  &CV-2D &CV-3D &MMVP &NaB.\\
    \hline

    $[0,1]$ & 65.0 & 335 & 30.8 & 61.5 & & 60.7 & 53.8 & 25.3 & 10.8 \\
    $[0, 0.5]$ & 67.8 & 349 & 32.5 & 62.5 & & 62.1 & 65.4 & 22.7 & 9.7 \\
    $[0.05,1]$ & 74.6 & 355 & 34.1 & 72.4 & & 59.9 & 62.3 & 41.3 & 18.3 \\
    $[0.05, 0.5]$ & 75.8 & 345 & 34.7 & 72.2 & & 63.5 & 68.7 & 38.7 & 20.5 \\
    \bottomrule
    \end{tabular}
    }
    \caption{Performance difference when applying different weights $[\alpha, \beta]$ for clamping.}
    \label{tab:ablation_cmo_hyp}
\end{table*}
\begin{table*}[!th]
    \centering
    \scalebox{0.8}{
    \begin{tabular}{cc|cccc c cccc}
    \toprule
    \multirow{2}*{Scales} &\multirow{2}{*}{Method} & \multicolumn{4}{c}{\textbf{General}} & &\multicolumn{4}{c}{\textbf{Vision-centric}}\\
    \cmidrule{3-6} \cmidrule{8-11}
    &&MMB. &OCRB. &DocVQA &AI2D &  &CV-2D &CV-3D &MMVP &NaB.\\
    \hline
    \multirow{2}{*}{1.5B} & Vanilla & 68.8 & 300 & 29.4 & 64.8 & & 55.9 & 54.5 & 22.7 & 12.8 \\
        & DPA & 70.7 & 321 & 30.6 &  65.6 & &57.9 & 61.1 & 34.7 & 14.3 \\
    \hline
    \multirow{2}{*}{3B} & Vanilla & 71.0 & 328 & 31.9 & 70.2&  & 57.5 & 66.5 & 31.3 & 14.4 \\
        & DPA & 72.6 & 354 & 34.9 &  70.3 && 56.8 & 67.9 & 31.3 & 17.1 \\
    \hline
    \multirow{2}{*}{14B} & Vanilla & 76.6 & 351 & 35.9 & 75.0 & & 63.3 & 64.9 & 38.7 & 18.5 \\
        & DPA & 78.4 & 351 & 33.9 & 73.5 & & 67.0 & 69.8 & 39.3 & 21.3 \\
    \hline
    \multirow{2}{*}{32B} & Vanilla & 79.8 & 369 & 38.1 & 78.0 & & 71.5 & 75.8 & 48.2 & 22.7 \\
        & DPA & 81.0 & 366 & 37.5  & 76.7 & & 71.8 & 75.9 & 54.4 & 22.9 \\
    \bottomrule
    \end{tabular}
    }
    \caption{Performance difference across different LLM parameter scales.}
    \label{tab:scaling_llm}
\end{table*}
\begin{table*}[!th]
    \centering
    \scalebox{0.8}{
    \begin{tabular}{cc|cccc c cccc}
    \toprule
    \multirow{2}*{Data size} &\multirow{2}{*}{Method} & \multicolumn{4}{c}{\textbf{General}} & &\multicolumn{4}{c}{\textbf{Vision-centric}}\\
    \cmidrule{3-6} \cmidrule{8-11}
    &&MMB. &OCRB. &DocVQA &AI2D &  &CV-2D &CV-3D &MMVP &NaB.\\
    \hline
    \multirow{2}{*}{5\%} & Vanilla & 44.2 & 27 & 0.4 & 56.6 & & 4.6 & 0.6 & 0.7 & 1.2 \\
        & DPA & 19.0 & 22 & 0.3 & 57.3 & & 1.6 & 0.0 & 0.0 & 0.0 \\
    \hline
    \multirow{2}{*}{10\%} & Vanilla & 47.9 & 59 & 0.4 & 53.8 & & 6.3 & 0.0 & 0.0 & 2.1 \\
        & DPA & 46.7 & 74 & 0.3 & 52.8 & & 0.4 & 0.0 & 0.0 & 2.3\\
    \hline
    \multirow{2}{*}{25\%} & Vanilla & 53.8 & 158 & 11.4 & 49.7 & & 53.8 & 55.8 & 20.7 & 10.9 \\
        & DPA & 52.4 & 276 & 19.0 & 46.9 & & 57.9 & 61.2 & 38.7 & 16.7 \\
    \hline
    \multirow{2}{*}{50\%} & Vanilla & 71.8 & 255 & 16.8 & 64.8 & & 56.0 & 68.2 & 31.3 & 16.4 \\
        & DPA & 75.4 & 256 & 18.0 & 64.9 & & 61.2 & 70.8 & 37.3 & 17.5 \\
    \hline
    \multirow{2}{*}{100\%} & Vanilla & 73.1 & 365 & 34.7 & 72.9 & & 58.5 & 62.1 & 36.0 & 18.5\\
        & DPA & 76.1 & 368 & 36.0 & 74.1 & & 61.1 & 68.1 & 40.7 & 20.4\\
    \bottomrule
    \end{tabular}
    }
    \caption{Performance difference across different data size.}
    \label{tab:scaling_data}
\end{table*}
\begin{table*}[!th]
    \centering
    \scalebox{0.8}{
    \begin{tabular}{lccccc c cccc}
    \toprule
    \multirow{2}*{Method}&  \multirow{2}*{LoRA rank} & \multicolumn{4}{c}{\textbf{General}} & &\multicolumn{4}{c}{\textbf{Vision-centric}}\\
    \cmidrule{3-6} \cmidrule{8-11}
    & &MMB. &OCRB. &DocVQA &AI2D &  &CV-2D &CV-3D &MMVP &NaB.\\
    \hline
    DPA w/o LoRA & -  & 73.7 & 360 & 34.6 & 72.4 &  & 59.3 & 67.2 & 34.0 & 17.6\\
    \hline
    \multirow{3}{*}{DPA with LoRA} &128 &73.0 & 343 & 32.7 & 73.2 &  & 60.6 & 67.1 & 28.7 & 17.6 \\
    &256 &75.0 & 350 & 33.6 & 73.2 &  & 60.1 & 67.5 & 37.3 & 18.0 \\
    &512 &74.1 & 348 & 33.8 & 72.5 &  & 58.6 & 65.4 & 35.3 & 18.1 \\
    \bottomrule
    \end{tabular}
    }
    \caption{Performance difference when applying different rank for LoRA. The LoRA alpha is set to twice the LoRA rank.}
    \label{tab:ablation_rank}
\end{table*}
\begin{table*}[!th]
    \centering
    \scalebox{0.8}{
    \begin{tabular}{lcccc c cccc}
    \toprule
    \multirow{2}*{Initialization} & \multicolumn{4}{c}{\textbf{General}} & &\multicolumn{4}{c}{\textbf{Vision-centric}}\\
    \cmidrule{2-5} \cmidrule{7-10}
    &MMB. &OCRB. &DocVQA &AI2D &  &CV-2D &CV-3D &MMVP &NaB.\\
    \hline

    Proxy LLM& 75.1 & 363 & 34.0 & 73.2 & & 61.9 & 67.9 & 38.7 & 18.6 \\
    Target LLM & 75.8 & 345 & 34.7 & 72.2 & & 63.5 & 68.7 & 38.7 & 20.5 \\
    \bottomrule
    \end{tabular}
    }
    \caption{Performance comparison between models initialized with the proxy LLM and the target LLM for instruction fine-tuning.}
    \label{tab:ablation_replace}
\end{table*}
\end{document}